\documentclass{article}

\PassOptionsToPackage{numbers}{natbib}
\usepackage[preprint]{neurips_2024}

\usepackage[utf8]{inputenc} 
\usepackage[T1]{fontenc}    
\usepackage{hyperref}       
\usepackage{url}            
\usepackage{booktabs}       
\usepackage{amsfonts}       
\usepackage{nicefrac}       
\usepackage{microtype}      
\usepackage{xcolor}         
\usepackage{graphicx}
\usepackage{placeins}
\usepackage{calc}
\usepackage{algorithm}
\usepackage{algpseudocode}
\usepackage{listings}
\usepackage{amsmath}
\usepackage{amsthm}
\usepackage{amssymb}
\usepackage{algorithm}
\usepackage{multirow}
\usepackage[numbers]{natbib}
\newtheorem{theorem}{Theorem}[section]
\newcommand{\code}[1]{\lstinline|#1|}

\title{Attention-Driven Metapath Encoding in Heterogeneous Graphs}

\makeatletter
\renewcommand{\@noticestring}{}
\makeatother
\author{%
  Calder K. Katyal\\
  Undergraduate\\
  Yale University\\
  New Haven, CT 06511 \\
  \texttt{calder.katyal@yale.edu} \\
}
\begin{document}

\maketitle

\begin{abstract}

One of the emerging techniques in node classification in heterogeneous graphs is to restrict message aggregation to pre-defined, semantically meaningful structures called metapaths. This work is the first attempt to incorporate attention into the process of encoding entire metapaths without dropping intermediate nodes. In particular, we construct two encoders: the first uses sequential attention to extend the multi-hop message passing algorithm designed in \citet{magna} to the metapath setting, and the second incorporates direct attention to extract semantic relations in the metapath. The model then employs the intra-metapath and inter-metapath aggregation mechanisms of \citet{han}. We furthermore use the powerful training scheduler specialized for heterogeneous graphs that was developed in \citet{lts}, ensuring the model slowly learns how to classify the most difficult nodes. The result is a resilient, general-purpose framework for capturing semantic structures in heterogeneous graphs. In particular, we demonstrate that our model is competitive with state-of-the-art models on performing node classification on the IMDB dataset, a popular benchmark introduced in \citet{benchmark}.

\end{abstract}

\section{Background}

Since the invention of Graph Attention Networks (GATs) in \citet{gat}, a variety of approaches have sought to extend the attention mechanism to heterogeneous domains. The GAT fails to account for the semantic relationships introduced by varying node and edge types, so a mechanism that incorporates this information into its base architecture is required. The heterogeneous attention network (HAN) in \citet{han} was the first method to incorporate semantic information into the attention mechanism by combining node-level and semantic-level attentions. In particular it employs \textbf{metapaths}, where a metapath $\Phi_j$ is formally defined as a sequence of nodes and edges in the form $V_1 \rightarrow{E_1} V_2 \rightarrow{E_2} \cdots \rightarrow{E_l} V_{l+1}$ that represents a composite relation between nodes $V_1$ and $V_{l+1}$. For example, in the IMDB dataset a particular metapath may take the form of ['Movie', 'Actor', 'Movie']. A \textbf{metapath instance} is a particular realization of that pattern (for example, {\textit{Inception},\textit{Leonardo DiCaprio}, \textit{Titanic}\}). We define the $\textbf{metapath-based neighbors}$ of a node $v_i$ and a metapath $\Phi_j$ as the set of all nodes $v_j$ which are terminal nodes of any instance of $\Phi$ starting at $v_i$. For each $v_i$ and $\Phi_j \in \{\Phi_1, \Phi_2, ..., \Phi_P\}$, HAN first calculates node-level attention by applying a node type specific projection so that all nodes share a feature space, and (2) aggregating the result of  attending to each metapath-based neighbor for each metapath instance of $\Phi_j$ (discarding non-terminal "intermediate" nodes) to obtain $P$ groups of semantic-specific embeddings. It then calculates semantic-level attention by transforming the semantic embeddings via a MLP and measuring similarity with a learnable semantic-level attention vector $\mathbf{q}$. This step measures the importance of each metapath. 

Addressing (2), \citet{magnn} incorporates intermediate nodes by employing intra-metapath aggregation. Specifically, for each $v_i$ and metapath $\Phi_j$, they use a metapath instance encoder to separately transform the features of all of the nodes in each metapath instance originating at $v_i$ into a single vector. Encoder options include mean, linear, and relational rotation (\citet{rotatee}). Attention is then used to create a weighted sum of all of the metapath instances of $\Phi_j$ that contain $v_i$. Unlike HAN, \citet{sehgnn} proposes a model which completely removes node-level attention, replacing it with simple aggregation. Other methods choose to not employ metapaths at all.  It is shown in \citet{lts} that many of these methods can be augmented by a Loss-aware Training Schedule (LTS). The scheduler first prioritizes the easiest nodes during training and gradually introduces the more difficult nodes to mitigate the impact of noisy data. 

A completely different approach is found in \citet{magna}. The model in the paper, Multi-hop Attention Graph Neural Network (MAGNA), uses a diffusion mechanism to have nodes multiple hops away attend to each other in each layer. In particular, it supports the heterogeneous graph setting by incorporating relation embeddings. However, it does not exploit long-range relations (e.g. metapaths). This is the main subject of our work.

\section{Method}

We construct two novel metapath instance encoders. The goal is to create a meaningful embedding for each instance such that they can be meaningfully aggregated using the simple attention-based method of HAN as in \citet{han}. Our codebase is public; we have also implemented the base model HAN for ablation in PyTorch Geometric.\footnote{https://github.com/calderkatyal/CPSC483FinalProject}

\subsection{Metapath Extraction}

As in \citet{magnn}, we seek to incorporate intermediate nodes into our metapaths. We develop a simple but novel method to do so. In particular, we modify \texttt{transforms.AddMetaPaths} to store a dictionary of the specific node types and node IDs in each metapath instance of each metapath. We store the features of the original graph so we can easily access the features in each metapath before performing the metapath transform, which constructs edges between terminal nodes in each metapath instance and removes intermediate nodes.

\subsection{Multi-hop Encoder}

Our first encoder extends the multi-hop framework MAGNA in \citet{magna} as to metapaths, where we use all node features in the metapath (which we stored while extracting metapaths). Recall that MAGNA first calculates one-hop attention scores via:
\begin{equation}
s_{i,k,j}^{(l)} = \delta(v_a^{(l)} \tanh(W_h^{(l)}h_i^{(l)} \| W_t^{(l)}h_j^{(l)} \| W_r^{(l)}r_k))
\end{equation}
where $\delta = \text{LeakyReLU}$, $W_h^{(l)}, W_t^{(l)} \in \mathbb{R}^{d^{(l)} \times d^{(l)}}$, 
$W_r^{(l)} \in \mathbb{R}^{d^{(l)} \times d_r}$ and $v_a^{(l)} \in \mathbb{R}^{1 \times 3d^{(l)}}$ are trainable weights. 
The attention score matrix $S^{(l)}$ is then computed as:
\begin{equation}
S_{i,j}^{(l)} = \begin{cases} 
s_{i,k,j}^{(l)}, & \text{if } (v_i, r_k, v_j) \text{ appears in } \mathcal{G} \\ 
-\infty, & \text{otherwise} 
\end{cases}
\end{equation}
Finally, the attention matrix $A^{(l)}$ is obtained by performing row-wise softmax over $S^{(l)}$:
\begin{equation}
A^{(l)} = \text{softmax}(S^{(l)})
\end{equation}
where $A_{i,j}^{(l)}$ denotes the attention value at layer $l$ when aggregating message from node $j$ to node $i$. Next, MAGNA builds a multi-hop attention diffusion matrix:
\begin{equation}
\mathcal{A} = \sum_{m=0}^{\infty} \gamma(1-\gamma)^m A^{(l-1)^m}, 0 < \gamma < 1
\end{equation}
The node embeddings are then updated by aggregating messages based on this multi-hop attention:
\begin{equation}
h_i^{(l)} = \sum_{j \in N_i} \mathcal{A}_{ij}h_j^{(l-1)}
\end{equation}
where $N_i$ represents the set of multi-hop neighbors.

We now show that these calculations simplify significantly when restricted to metapaths and there emerges good interpretability. Consider a metapath instance $\Phi$ of length $|\Phi| - 1 = k$ with vertices $\{v_0, v_1, v_2, \dots, v_k\}$. We seek to learn an embedding $(\mathbf{h}_0)^\prime$ for the source node. The key is realizing that we can think of $\Phi$ as a directed chain $v_0 \leftarrow v_1 \leftarrow v_2 \leftarrow \dots \leftarrow v_k$. This structure makes intuitive sense as we would expect the nearest neighbor $v_1$ of $v_0$ to provide a more informative message than the far away node $v_k$. It also makes sense that the feature for the source $\mathbf{h}_0$ should influence the updated embedding; this is the same as adding a virtual self-loop around the source. Let $\alpha_{ij} = A_{ij}$ denote attention score (e.g. how much $v_i$ "attends to" or "is influenced by" $v_j$). Due to the virtual self-loop, $\mathbf{A}^0 = \text{diag}(\alpha_{ii})_{i=0}^{k}$. The following result follows:

\begin{theorem}
For a metapath instance $\Phi$ of length $k$ with vertices $\{v_0, v_1, \dots, v_k\}$ with a self-loop at the source, the updated embedding $(\mathbf{h}_0)^\prime$ for the source node given by MAGNA (with a single layer) is given by:
\begin{equation}
    \begin{aligned}
        (\mathbf{h}_0)^\prime &= \gamma \mathbf{h}_0 a_{00} + \sum_{i=1}^k \gamma(1-\gamma)^i \mathbf{h}_i \prod_{j=1}^i a_{j(j-1)} \\
        &= \gamma \mathbf{h}_0 a_{00} + \gamma(1-\gamma) \mathbf{h}_1 a_{10} + \gamma(1-\gamma)^2 \mathbf{h}_2 (a_{21} a_{10}) \\
        &\quad + \gamma(1-\gamma)^3 \mathbf{h}_3 (a_{32} a_{21} a_{10}) + \ldots + \gamma(1-\gamma)^k \mathbf{h}_k (a_{k(k-1)} \ldots a_{21} a_{10}).
    \end{aligned}
\end{equation}
\end{theorem}

The proof is given in Appendix \ref{proof1}.  As shown in \citet{magna}, multihop attention is equivalent to putting a Personalized Page Rank (PPR) prior on attention values. Thus we get good interpretability for our metapath encodings; not only do far away nodes contribute less to the encoding than closer nodes, information flows in a semantically meaningful way. For instance, consider the metapath instance \{\textit{Inception},\textit{Leonardo DiCaprio}, \textit{Titanic}\}, where the goal is to get a good embedding for Inception for movie genre classification. Intuitively, the genre of \textit{Inception} is influenced directly by the fact that Leonardo Di Caprio starred in it and also influenced indirectly by the movie \textit{Titanic}, mediated through the intermediary of Leonardo DiCaprio himself (in other words, the semantic relationship between the two movies exists only through his shared involvement in both).

Therefore, it suffices to compute attention between all pairs $(v_i, {v_{i-1}})_{i=1}^k$ and self-attention for the source. To do so, we use a 1-layer version of MAGNA's algorithm, employing (1) without relationship embeddings and (3) with sigmoid instead of softmax (a simplex constraint for attention weights is too strong for structures with such semantic richness). We then use (6) to get metapath instance encodings (a $\gamma$ value of 0.4 seems to work well in practice). 

\subsection{Direct Attention Encoder}

Our second encoder is a simplified version that intuitively works well for short metapaths (which are common in practice). Let $\Phi = \{v_0, v_1, v_2, \dots, v_k\}$ be a metapath instance. We now compute the embedding $\mathbf{h}_0^\prime$ by directly attending to all nodes in the metapath.

First, features are transformed using learnable matrices $\mathbf{W}_t$ for the source node $v_0$ and $\mathbf{W}_h$ for all other nodes:
\begin{equation}
\mathbf{h}_0 \triangleq \mathbf{h}_0 \mathbf{W}_t, \quad 
\mathbf{h}_i \triangleq \mathbf{h}_{\text{raw}, i} \mathbf{W}_h, \quad i \in \{1, 2, \dots, k\}.
\end{equation}

Attention scores $\alpha_{0i}$ are computed between $v_0$ and each node $v_i$ (including self-attention for $v_0$) as:
\begin{equation}
\alpha_{i0} = \text{sigmoid}\left(\frac{\langle \mathbf{h}_0, \mathbf{h}_i \rangle}{\sqrt{d}}\right), \quad i \in \{0, 1, 2, \dots, k\}.
\end{equation}

The final embedding for the source node $v_0$ is then computed as a weighted sum of all transformed node features:
\begin{equation}
\mathbf{h}_0^\prime = \sum_{i=0}^k \alpha_{i0} \mathbf{h}_i.
\end{equation}

This encoder allows the source node to perform weighted aggregation on all node features in the metapath at the cost of lesser interpretability for longer metapaths (where it is vital that distant nodes attend to each other in a controlled manner). 

\subsection{Algorithm}

We now give the full algorithm for our model HAN-ME (HAN with Metapath Encoding). We first perform a type-specific projection to ensure all node features share a common feature space as HAN in \citet{han}. We then diverge from HAN by encoding each metapath instance using either multi-hop or direct attention. To aggregate over metapath instances, we use attention between the embedding of the source node and the metapath instance encoding. Inter-metapath aggregation is performed the same as in HAN. Our loss function is binary cross-entropy with logits.

\begin{algorithm}[ht]
\caption{HAN-ME: Heterogeneous Attention Network with Metapath Encoding}\label{algorithm1}
\small
\begin{algorithmic}[1]
    \State \textbf{Input:} 
    \Statex \hspace{2em} The heterogeneous graph $\mathcal{G} = (\mathcal{V}, \mathcal{E})$; the node feature $\{\mathbf{h}_i, \forall i \in \mathcal{V}\}$
    \Statex \hspace{2em} The metapath set $\{\Phi_0, \Phi_1, \ldots, \Phi_P\}$; the number of attention heads $K$
    \State \textbf{Output:} 
    \Statex \hspace{2em} The final embedding $\mathbf{Z}$
    \For{$\Phi_i \in \{\Phi_0, \Phi_1, \ldots, \Phi_P\}$}
        \For{$k = 1\ldots K$}
            \State Type-specific transformation: $\mathbf{h}_i \leftarrow \mathbf{M}_{\phi_i} \cdot \mathbf{h}_i$ \Comment{$\mathbf{M}_{\phi_i}$ is transformation matrix for node type $\phi_i$}
            \For{$i \in \mathcal{V}$}
                \For{each metapath instance $\Phi_{ij}$ rooted at $v_i$}
                    \State Metapath Encoding: $h_{ij} \leftarrow \text{Encoder}(\Phi_{ij})$ \Comment{Multihop or direct attention}
                    \State Calculate importance: $e_{ij}^\Phi = \text{att}_{\text{node}}(\mathbf{h}_i, \mathbf{h}_{ij}; \Phi_{ij})$ \Comment{Node-level attention score}
                    \State Normalize weights: $\alpha_{ij}^\Phi = \frac{\exp(\sigma(\mathbf{a}_\Phi^\top \cdot [\mathbf{h}_i \| \mathbf{h}_{ij}]))}{\sum_{k \in \mathcal{N}_i^\Phi} \exp(\sigma(\mathbf{a}_\Phi^\top \cdot [\mathbf{h}_i \| \mathbf{h}_{ik}]))}$ \Comment{$\mathbf{a}_\Phi$ is attention vector}
                \EndFor
                \State Calculate semantic-specific node embedding:
                \State $\mathbf{z}_i^\Phi \leftarrow \sigma(\sum_{j \in \mathcal{N}_i^\Phi} \alpha_{ij}^\Phi \cdot \mathbf{h}_j')$ \Comment{Aggregate neighbor features}
            \EndFor
            \State Concatenate learned embeddings from all attention heads:
            \State $\mathbf{z}_i^\Phi \leftarrow \|_{k=1}^K \sigma(\sum_{j \in \mathcal{N}_i^\Phi} \alpha_{ij}^\Phi \cdot \mathbf{h}_j')$ 
        \EndFor
        \State Calculate metapath importance:
        \State $w_{\Phi_i} = \frac{1}{|\mathcal{V}|} \sum_{i \in \mathcal{V}} \mathbf{q}^\top \cdot \tanh(\mathbf{W} \cdot \mathbf{z}_i^\Phi + \mathbf{b})$ \Comment{$\mathbf{q}$ is semantic attention vector}
        \State Normalize metapath weights:
        \State $\beta_{\Phi_i} = \frac{\exp(w_{\Phi_i})}{\sum_{i=1}^P \exp(w_{\Phi_i})}$ \Comment{Softmax over metapaths}
        \State Fuse semantic-specific embedding:
        \State $\mathbf{Z} \leftarrow \sum_{i=1}^P \beta_{\Phi_i} \cdot \mathbf{Z}_{\Phi_i}$ \Comment{Weighted combination of metapaths}
    \EndFor
    \State Calculate Binary Cross-Entropy with Logits:
\State $L = -\sum_{l \in \mathcal{Y}_L} \mathbf{Y}_l \log(\sigma(\mathbf{Z}_l)) + (1-\mathbf{Y}_l)\log(1-\sigma(\mathbf{Z}_l))$ \Comment{Optionally use LTS to find lowest loss nodes}
    \State Back propagation and update parameters in HAN-ME
    \State \Return{$\mathbf{Z}$}
\end{algorithmic}
\end{algorithm}

\section{Experiments}

We use the multi-label IMDB dataset introduced by \citet{imdb} for heterogeneous node classification. The dataset is one of the most common for this task and is known for its noisiness and difficulty. The dataset comprises 21,420 nodes (4,932 movies, 2,393 directors, 6,124 actors, 7,971 keywords) with 86,642 bidirectional edges for each relation type. Only movie nodes contain sparse features, such as "budget," "duration," and "language." They are labeled with up to five genres (Drama: 2,517, Comedy: 1,837, Thriller: 1,384, Action: 1,135, Romance: 1,090), with most having one or two labels. We split the data into training (1,096 nodes, 22.22\%), validation (275 nodes, 5.58\%), and test (3,202 nodes, 64.92\%) sets, omitting unlabeled nodes as is standard practice with this benchmark. We perform minimal preprocessing except for assigning features to director, actor, and keyword nodes through mean pooling of the connected node features. As in \citet{han}, we employ the metapaths \{\textit{Movie}, \textit{Director}, \textit{Movie}\} and \{\textit{Movie}, \textit{Actor}, \textit{Movie}\}. 

To train our model we use a learning rate of 0.005, weight decay of 0.001, a dropout rate of 0.6, 8 attention heads, and 128 hidden units as in \citet{han}, as well as a random seed of 483 and a patience level of 100 for early stopping. Furthermore, $\gamma=0.4$ seems to work well in practice for the teleportation rate in the multihop encoder. 

We also implement LTS as in \citet{lts}. LTS ranks nodes by their loss values during training and progressively introduces them from easiest to hardest, where at epoch $t$ a proportion $\lambda_t$ of nodes are included according to one of three pacing functions: linear ($\lambda_t = \lambda_0 + (1-\lambda_0)\frac{t}{T}$), root ($\lambda_t = \sqrt{\lambda_0^2 + (1-\lambda_0^2)\frac{t}{T}}$), or geometric ($\lambda_t = 2^{\log_2(\lambda_0) - \log_2(\lambda_0)\frac{t}{T}}$), with $\lambda_0$ being the initial proportion and $T$ the epoch when all nodes are included. For our results we use $\lambda_0 = 0.1$ and the linear scheduler. We select the model with highest validation performance as our final model. Training performance is summarized below: 

\begin{table}[ht]
  \caption{Performance of HAN-ME Multihop and HAN-ME Direct With Baseline HAN}
  \label{model-comparison}
  \centering
  \small
  \begin{tabular}{lcccc}
    \toprule
    \multirow{2}{*}{Model} & \multicolumn{2}{c}{Micro F1} & \multicolumn{2}{c}{Macro F1} \\
    \cmidrule(r){2-3} \cmidrule(r){4-5}
    & Standard & With LTS & Standard & With LTS \\
    \midrule
    HAN-ME Multihop & $\mathbf{0.6801 \pm 0.0001}$ &  $0.6737 \pm 0.0005$ & $0.6353 \pm 0.0008$ & $0.6222 \pm0. 0005 $ \\
    HAN-ME Direct & $0.6767 \pm 0.0006$ & $0.6769\pm 0.0003$ & $\mathbf{0.6418 \pm 0.0007}$& $ 0.6380\pm 0.0004$\\
    HAN & $0.6426 \pm 0.0006$ & $ 0.6427\pm 0.0007$ & $0.5949 \pm 0.0010$ & $0.5966\pm 0.0009$ \\
    \bottomrule
  \end{tabular}
\end{table}

\begin{figure}[ht]
\label{image1}
  \centering
  \includegraphics[width=\textwidth]{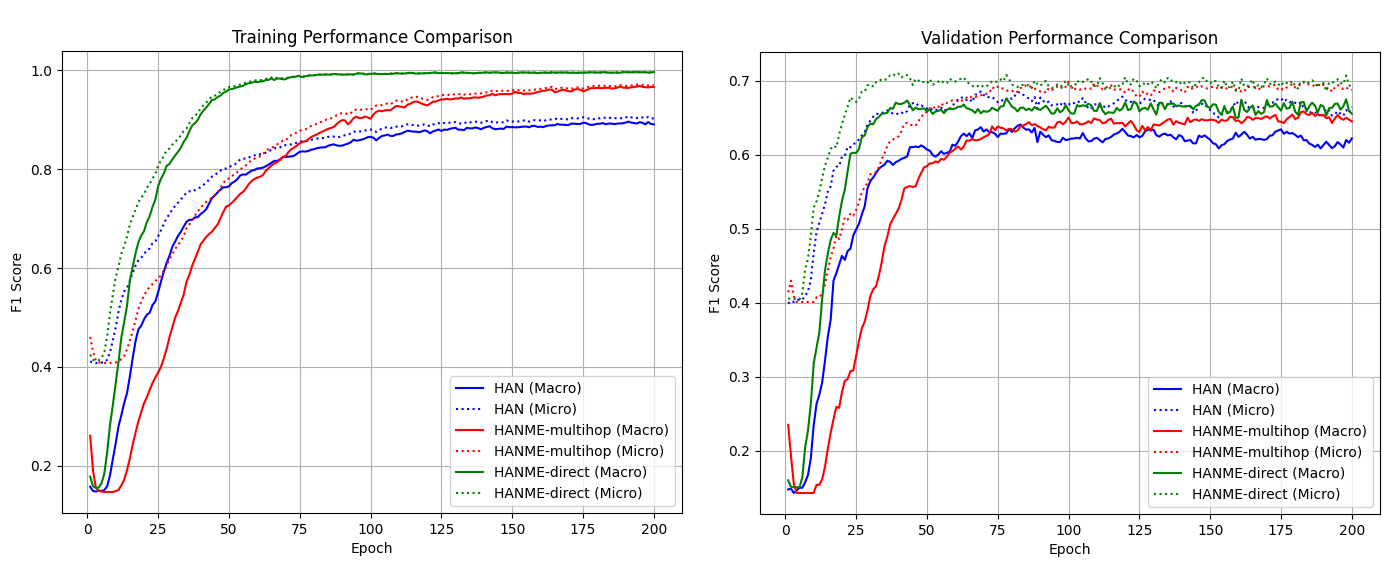}
  \caption{Micro F1 and MacroF1 on Training and Validation Sets}
\end{figure}

We see that the multihop encoder yields the best test Micro F1 and the direct attention encoder yields the best test Macro F1. Notably, while the two perform significantly better than HAN, they perform at around the same level. This is likely due to two factors (1) We only use metapaths with 3 nodes, for which the direct attention encoder has good interpretability, and (2) We did not have explicit node features for non-movie nodes, which would significantly improve the multihop method. Additionally, LTS (which we tested with multiple configurations) did not cause a statistically significant difference in performance. This islikely because the dataset features many very difficult nodes which require significant message passing and neighbor information to classify correctly. As seen in \ref{image1}, the models overfit to the training data (as expected) and suffer somewhat in generalization. 

\section{Conclusion}

In this paper we introduced HAN-ME, a novel framework for incorporating attention-based metapath instance encoding into heterogeneous graph neural networks. We presented two distinct encoding approaches: a multi-hop encoder that extends MAGNA's diffusion mechanism to metapaths and a direct attention encoder. Both encoders show significant improvements over the baseline HAN model. We believe the multihop method, with its strong interpretability, has the potential to deliver better performance if applied to longer metapaths or if specific components, such as its activation function, were optimized. Our experimental results provide compelling evidence for the effectiveness of HAN-ME in capturing complex heterogeneous graph structures.

\newpage


\appendix

\section*{Appendix}

\section{Proof of Theorem 1}
\label{proof1}

We prove (6). Due to the virtual self-loop, we have that $\mathbf{A}^0 = \text{diag}(\alpha_{ii})_{i=0}^{k}$. Then:
\[
\mathbf{A}^1 =
\begin{bmatrix}
0 & \alpha_{01} & 0 & \cdots & 0 & 0 & 0 \\
0 & 0 & \alpha_{12} & \cdots & 0 & 0 & 0 \\
0 & 0 & 0 & \ddots & 0 & 0 & 0 \\
\vdots & \vdots & \vdots & \ddots & \alpha_{k-2,k-1} & 0 & 0 \\
0 & 0 & 0 & \cdots & 0 & \alpha_{k-1,k} & 0 \\
0 & 0 & 0 & \cdots & 0 & 0 & \alpha_{k,k+1} \\
0 & 0 & 0 & \cdots & 0 & 0 & 0
\end{bmatrix},
\]

\[
\mathbf{A}^2 = 
\begin{bmatrix}
0 & 0 & \alpha_{01}\alpha_{12} & \cdots & 0 & 0 & 0 \\
0 & 0 & 0 & \ddots & 0 & 0 & 0 \\
0 & 0 & 0 & \ddots & \alpha_{(k-3)(k-2)}\alpha_{(k-2)(k-1)} & 0 & 0 \\
\vdots & \vdots & \vdots & \ddots & 0 & \alpha_{(k-2)(k-1)}\alpha_{(k-1)k} & 0 \\
0 & 0 & 0 & \cdots & 0 & 0 & \alpha_{(k-1)k}\alpha_{k(k+1)} \\
0 & 0 & 0 & \cdots & 0 & 0 & 0 \\
0 & 0 & 0 & \cdots & 0 & 0 & 0
\end{bmatrix},
\]

\[
\mathbf{A}^{k-1} =
\begin{bmatrix}
0 & 0 & 0 & \cdots & 0 & 0 & \alpha_{01}\alpha_{12}\cdots\alpha_{(k-1)(k)}\alpha_{(k)(k+1)} \\
0 & 0 & 0 & \cdots & 0 & 0 & 0 \\
0 & 0 & 0 & \cdots & 0 & 0 & 0 \\
\vdots & \vdots & \vdots & \ddots & \vdots & \vdots & \vdots \\
0 & 0 & 0 & \cdots & 0 & 0 & 0 \\
0 & 0 & 0 & \cdots & 0 & 0 & 0 \\
0 & 0 & 0 & \cdots & 0 & 0 & 0
\end{bmatrix}
\]

Note that $A$ is nilpotent (e.g. $\forall p \geq k$, $\mathbf{A}^p = \mathbf{0}$). Therefore (4) becomes
\[
\mathcal{A} = \sum_{m=0}^\infty \gamma (1 - \gamma)^k \mathbf{A}^k = \sum_{m=0}^{k} \gamma (1 - \gamma)^m \mathbf{A}^m,
\]

For a metapath instance \( \Phi \) with vertices \( \{v_0, v_1, v_2, \dots, v_k\} \), the entry \( (\mathcal{A})_{0j} \) represents the attention contribution from node \( v_j \) to \( v_0 \). This is computed as:
\[
(\mathcal{A})_{0j} = \sum_{i=0}^k \gamma (1 - \gamma)^i \prod_{t=1}^i a_{t(t-1)},
\]

The embedding \( \mathbf{h}_0^\prime \) for the source node is then updated by aggregating messages from all other nodes \( v_j \) in the metapath instance. Using the definition of \( \mathcal{A} \), we have:
\[
\mathbf{h}_0^\prime = \sum_{j=0}^k (\mathcal{A})_{0j} \mathbf{h}_j,
\]
which expands to:

\begin{equation*}
    \begin{aligned}
        (\mathbf{h}_0)^\prime &= \gamma \mathbf{h}_0 a_{00} + \sum_{i=1}^k \gamma(1-\gamma)^i \mathbf{h}_i \prod_{j=1}^i a_{j(j-1)} \\
        &= \gamma \mathbf{h}_0 a_{00} + \gamma(1-\gamma) \mathbf{h}_1 a_{10} + \gamma(1-\gamma)^2 \mathbf{h}_2 (a_{21} a_{10}) \\
        &\quad + \gamma(1-\gamma)^3 \mathbf{h}_3 (a_{32} a_{21} a_{10}) + \ldots + \gamma(1-\gamma)^k \mathbf{h}_k (a_{k(k-1)} \ldots a_{21} a_{10}).
    \end{aligned}
\end{equation*}


\end{document}